\documentclass[sigconf]{acmart}
\AtBeginDocument{%
  \providecommand\BibTeX{{%
    \normalfont B\kern-0.5em{\scshape i\kern-0.25em b}\kern-0.8em\TeX}}}

\setcopyright{acmcopyright}
\copyrightyear{2018}
\acmYear{2018}
\acmDOI{}

\acmConference[CBMI 2023]{ In Proceedings of
International Conference on Content-based Multimedia Indexing}{September 20--22,
  2023}{Orleans, France}
\acmPrice{15.00}
\acmISBN{978-1-4503-XXXX-X/18/06}

\usepackage{multirow}
\usepackage{booktabs}
\usepackage{graphics}
\usepackage{adjustbox}
\usepackage{subcaption}
\usepackage[font=small,labelfont=bf]{caption}
\usepackage{url}
\usepackage{graphicx}
\usepackage{array}

\begin{document}

\title{Explanation of Face Recognition via Saliency Maps}

\author{Yuhang Lu and Touradj Ebrahimi}
\affiliation{%
  \institution{\'Ecole Polytechnique F\'ed\'erale de Lausanne (EPFL)}
  \streetaddress{}
  \city{Lausanne}
  \country{Switzerland}
  \postcode{1015}
}
\email{firstname.lastname@epfl.ch}

% \author{Touradj Ebrahimi}
% \affiliation{%
%   \institution{\'Ecole Polytechnique F\'ed\'erale de Lausanne (EPFL)}
%   \streetaddress{}
%   \city{Lausanne}
%   \country{Switzerland}
%   \postcode{1015}
% }
% \email{touradj.ebrahimi@epfl.ch}

\renewcommand{\shortauthors}{Lu and Ebrahimi}

\begin{abstract}

Despite the significant progress in face recognition in the past years, they are often treated as ``black boxes'' and have been criticized for lacking explainability. It becomes increasingly important to understand the characteristics and decisions of deep face recognition systems to make them more acceptable to the public. 
Explainable face recognition (XFR) refers to the problem of interpreting why the recognition model matches a probe face with one identity over others. Recent studies have explored use of visual saliency maps as an explanation, but they often lack a deeper analysis in the context of face recognition. This paper starts by proposing a rigorous definition of explainable face recognition (XFR) which focuses on the decision-making process of the deep recognition model. Following the new definition, a similarity-based RISE algorithm (S-RISE) is then introduced to produce high-quality visual saliency maps. Furthermore, an evaluation approach is proposed to systematically validate the reliability and accuracy of general visual saliency-based XFR methods.

% This paper proposes a rigorous definition of explainable face recognition (XFR) which analyzes the decision-making process of the deep recognition model. Following the new definition of the explanation method, this study further introduces a similarity-based RISE algorithm (S-RISE) to interpret the face recognition process.  
% S-RISE is adapted and improved from the RISE algorithm by leveraging the similarity scores between two faces as weights for the masks and it provides explainable saliency maps without accessing the intrinsic architecture of the face recognition model.
% Furthermore, an evaluation approach is proposed to systematically validate the reliability and accuracy of general visual saliency-based XFR methods.

\end{abstract}






\maketitle

\section{Introduction}

Thanks to the rapid development of deep learning, recent years have witnessed a breakthrough in various computer vision tasks, such as image classification, object detection, and face recognition. Deep face recognition technology has attracted worldwide attention in the past decades due to its remarkable performance and wide applications in multiple areas, such as automatic border control, and security cameras, to mention a couple. Despite the beneficial usage, such biometric systems also have the potential to endanger fundamental privacy and data protection rights, raising serious public concerns. Besides, the ``black-box'' nature of the deep learning-based systems is another barrier to the deployment of  face recognition system based on the latter, which is often criticized for lacking interpretability.
Therefore, it is necessary to understand and to explain the decisions made by deep face recognition technologies in order to further improve their performance and make them more acceptable to society.

A number of explanation techniques were first proposed as forms of explainable artificial intelligence (XAI) to better understand AI-based models. For computer vision tasks, various visual saliency map algorithms \cite{binder2016layer,zhou2016learning,ribeiro2016should,selvaraju2017grad,chattopadhay2018grad,petsiuk2018rise,zhang2018top} have been introduced to highlight either the internal CNN layers or the important pixels of the input image that are relevant to the model's decision. 

Explainable face recognition (XFR) is the problem of explaining how the face recognition model verifies a given pair of faces. Although numerous visual saliency algorithms have achieved impressive results in classification tasks, they cannot be directly applied to other image-understanding tasks due to a notable difference in internal model structure and the output format. To address this issue, some studies have attempted to adapt the existing explanation method to the face recognition task. \cite{castanon2018visualizing,williford2020explainable} leveraged the contrastive excitation backpropagation (cEBP) technique \cite{zhang2018top} to localize important regions in the face. \cite{mery2022black} adopted a similar idea of perturbing input images as in \cite{ribeiro2016should,petsiuk2018rise}. \cite{winter2022demystifying} applied explainable boosting machine to face verification. 
Nevertheless, explaining a face recognition model not only refers to generating a saliency map but also involves an interpretation of why the model believes a certain pair of images is a better match than others and there is a lack of discussion regarding the latter.

This paper first presents a new definition of visual saliency-based explainable face recognition. Then a similarity-based RISE algorithm adapted from \cite{petsiuk2018rise} is introduced. It follows the new definition of XFR and provides insightful saliency explanations for the decision-making process of deep face recognition models. 

The contributions of this paper can be summarized as follows.
\begin{itemize}
    \item  A new definition for visual saliency map-based explainable face recognition is proposed. 
    \item A novel explanation method, called the Similarity-based RISE (S-RISE) algorithm, is proposed. Extensive visual explanation results have been presented, demonstrating its effectiveness.
    \item A new evaluation approach is introduced to quantitatively benchmark the saliency map-based XFR methods.
\end{itemize}

The remainder of this paper is organized as follows. A review of the state-of-the-art face recognition models and explainability methods is provided in section \ref{relatedwork}. The new proposed explanation method and evaluation approach are described in details in Section \ref{method}, followed by the experimental results in Section \ref{results} and the conclusion in Section \ref{conclusion}.

% D-RISE \cite{} adapted the RISE \cite{} algorithm for the object detection task. 
% similarity-based saliency algorithms \cite{} have been proposed for image retrieval tasks. 

% most of the current explanation methods focus on generating attention maps without further interpretation. 
% but also involves decision-making process of 
% Nevertheless, there is no such method that makes a comprehensive definition of what is 

% This paper explores new model-agnostic explainability approaches for deep face recognition systems. 

% Face verification refers to a typical scenario in face recognition, where a face verification system extracts deep features of a pair of face images to compute a similarity score in the embedding domain and compare it with a threshold value to decide whether they belong to the same subject. This paper mainly focuses on the problem of explainable face verification (XFV), which explains why a face recognition system 

% More specifically, a face verification system extracts deep features of a pair of face images to compute a similarity score in the embedding domain and compare it with a threshold value to decide whether they belong to the same subject. 

\section{Related Work}
\label{relatedwork}

\subsection{Face Recognition}
In recent years, deep learning has revolutionized the field of face recognition and current deep face recognition techniques have shown impressive performance on different public benchmarks. The advances of such technologies mainly come from the publicly available large-scale face datasets, powerful network architectures, and the evolution of training losses.

Early versions of deep face recognition systems were often built upon shallow network architectures, e.g. FaceNet \cite{schroff2015facenet} used GoogLeNet \cite{szegedy2015going} and VGGFace \cite{parkhi2015deep} used VGGNet \cite{simonyan2014very}. Then, ResNet \cite{simonyan2014very} introduced residual learning to train very deep networks and has become a popular network architecture for current face recognition models. \cite{hu2018squeeze} further improved the ResNet by introducing a squeeze and excitation module, often denoted by SE-ResNet.
While researchers first focused on investigating deeper backbone networks, the attention has gradually shifted to designing more discriminative loss functions. The most commonly used SoftMax loss \cite{liu2016large} was extended with an angular margin to improve the intra-class compactness and inter-class discrepancy. SphereFace \cite{liu2017sphereface} first applied a multiplicative angular margin in the face recognition task and obtained better performance than previous work. CosFace \cite{wang2018cosface} introduced easier supervision by directly adding a cosine margin penalty to the target logit. Subsequently, ArcFace \cite{deng2019arcface} proposed an additive angular margin loss to better separate the feature. More recently, the latest research, such as MagFace \cite{meng2021magface} and AdaFace \cite{kim2022adaface}, created loss functions that are adaptive to image quality. 

Despite the remarkable performance of current deep face recognition methods, they still suffer from some non-trivial scenarios according to recent  studies \cite{lu2022novel}, such as partial occlusion, low-resolution, and head pose variation.  It is necessary to have explainable tools that can provide insights into why a face recognition system fails in certain situations and how to take action to improve the current recognition process. 

\subsection{Explainable Artificial Intelligence (XAI)}

Explainable artificial intelligence is one of the most important problems in machine learning, which aims for a better comprehension of the decisions made by AI-based models, such as those by deep neural networks. There are several taxonomies of explanation methods based on their scope and mechanism. For example, a main criterion is \textit{intrinsic vs post-hoc}. Intrinsic methods refer to the machine learning models that are explainable by themselves, while post-hoc interpretability refers to explanation methods that are applied to a regular model after training. 
Explanations can also be either \textit{model-agnostic} or \textit{model-specific}. Model-specific explanation methods are often limited to specific model classes but the model-agnostic explanation can be applied to any AI model.  
This paper focuses on developing post-hoc and model-agnostic explanation methods, which interpret the decision of a deep model without accessing or modifying its intrinsic architecture. 

% proposes a post-hoc and model-agnostic explanation method.
% local methods interpret the decision of a black-box model for a specific set of samples, while global methods give an overall explanation of model behavior on any instance.

% that are applied to non-interpretable models. 

%%%%%%%%%%%%%%% S-RISE %%%%%%%%%%%%%%%%

\begin{figure*}[t]
	\centering
	\begin{adjustbox}{width=0.8\textwidth}
    \includegraphics[]{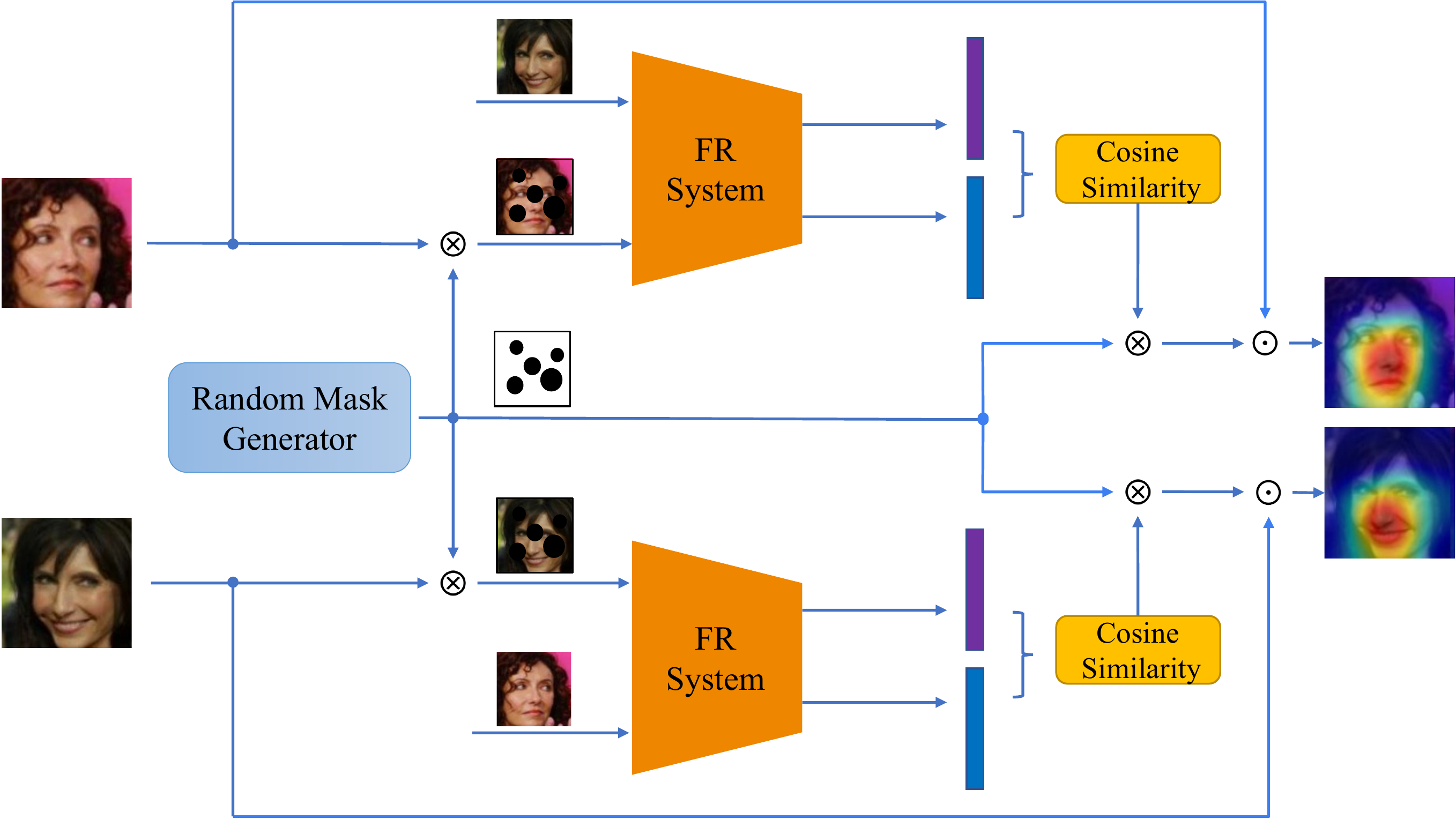}
	\end{adjustbox}
	\caption{Workflow of the proposed S-RISE explanation method.}
	\label{fig:srise}
\end{figure*}

%%%%%%%%%%%%%%%%%%%%%%%%%%%%%%%%%%%%%%%%%%%%%%%%%%%%%%%%%%

\subsection{Visual Saliency Methods}

Visual saliency algorithms have been widely employed to explain deep learning-based decision systems acting on images. A saliency map is itself an image, in which each pixel value represents the importance of the pixel. In practice, a saliency map is modeled as a 2-D matrix which is of the same size as the image to be explained. The value of each pixel in the saliency map represents the importance of the corresponding pixel in the input image. The saliency map provides insight into the important regions of the input that is responsible for the model's final decision. In general, there are two types of methods for creating saliency maps. 

The first group of methods backpropagate an importance score through the layers of the neural network from the model's output to the individual pixels in the input.  This type of approach often requires access to the intrinsic architecture or gradient information of the deep model. The class activation maps (CAM) method \cite{zhou2016learning} was one of the earliest work in this area,  replacing the fully-connected layers with a global average pooling layer and obtaining the class-specific importance region by computing a weighted sum of the feature activation values. Grad-CAM \cite{selvaraju2017grad} improved CAM by weighing the feature activation values with the class-specific gradient information that flows into the final convolutional layer of a CNN. Grad-CAM++ \cite{chattopadhay2018grad} further extended the Grad-CAM by providing a better visual explanation of CNN model predictions. Layer-wise Relevance Propagation (LRP) \cite{binder2016layer} provides post-hoc explanation by decomposition and is capable of interpreting the decisions of complex neural networks. Specifically, it redistributes the prediction of the neural network backward until it assigns a relevance score to each input image pixel.

The other type of methods perform random perturbations on the image, e.g. noise, occlusion, etc, and determine the importance region by observing the impact of such perturbation on the model's output prediction. For instance, Ribeiro et al. \cite{ribeiro2016should} proposed an interpretable approximate linear decision model (LIME) in the vicinity of a particular input which analyzes the relation between the input data and the prediction in a perturbation-based forward propagation manner. RISE \cite{petsiuk2018rise} algorithm applies random binary masks to the input image and then uses the output class probabilities as weights to compute a weighted sum of the masks as a saliency map.
While most of the popular visual saliency explanation methods are initially developed for  image classification tasks, there has been an increasing demand for creating explanations for other image-understanding tasks such as object detection \cite{fong2017interpretable, petsiuk2021black}.

In face recognition, earlier work \cite{castanon2018visualizing} adapted six saliency map creation methods for the face recognition task and compared their performance using a metric called the ``hiding game''.
\cite{yin2019towards} designed a feature activation diverse loss to encourage learning more interpretable face representations. 
\cite{williford2020explainable} first proposed a comprehensive benchmark for explainable face recognition, comprising two baseline explanation methods. Lin et al. \cite{lin2021xcos} proposed a learnable module that can be integrated into face recognition models and generates meaningful similarity maps. Mery et al. \cite{mery2022black} introduced six different perturbation-based methods to create saliency maps to explain the face verification model without manipulating the model. 

% RISE \cite{} algorithm applies random binary masks to the input image and then computes a weighted sum of the masks as the saliency map based on the output class probability. 
% For example, Vitali et al. \cite{} created a saliency map-based explanation for object detection tasks by leveraging the RISE algorithm. 
% Image retrieval, image similarity 
% \cite{zhong2019exploring} analyzed the attributes of VGGface to understand the mechanism. 

\section{Proposed Method}
\label{method}

\subsection{Definition for Explainable Face Recognition}
In the context of face recognition, the deep model predicts whether a pair of face images belongs to the same identity. Ideally, an explainable face recognition system gives a visual interpretation of why the model believes the given pair of faces is a match or non-match. Existing study \cite{williford2020explainable} has explored a similar idea of explaining how a deep face recognition model matches faces. It leverages a triplet of faces, i.e. probe, mate, and nonmate, to provide a deeper explanation for the relative importance of facial regions. Notably, probe refers to the query image to be verified, mate is the image from the same subject of the probe, and nonmate is the image from another subject.
More specifically,  explainable face recognition is defined as a way to highlight the regions of the probe image that can maximize the similarity to the mate image and meanwhile minimize the similarity to the nonmate. 

However, there is a drawback in this definition of explanation, as the most similar regions between the probe and the mate images are not necessarily the least similar regions between the probe and the nonmate. In fact, a face recognition system makes decisions by comparing a predefined threshold with the similarity score between two images instead of three, which means in the triplet, the decision-making process for each pair of images is independent.% of the third. 

This paper proposes a more rigorous definition of explainable face recognition, which preserves the idea of triplet images but disentangles the matching and non-matching pairs. Given a triplet $[probe, mate, nonmate]$ feeding into a face recognition system respectively, the explanation method should produce the corresponding saliency maps for the $[probe, mate]$ and $[probe, nonmate]$ pairs, which answer to the following questions. 
\begin{itemize}
    \item Which regions in the $[probe, mate]$ image pairs are the most similar to the FR system? 
    \item Which regions in the $[probe, nonmate]$ image pairs are the most similar to the FR system? 
    \item Why the FR system believes that $[probe, mate]$ pair is a better matching than $[probe, nonmate]$?
\end{itemize}

% the face recognition system only
% The face recognition system gives decisions by comparing a predefined threshold and the similarity of an image pair instead of a triplet. 

% First, the face recognition system gives verification results by comparing a predefined threshold and the similarity of an image pair instead of a triplet. 
% Secondly, the most similar regions between the probe and mate images are not necessarily the least similar regions between the probe and nonmate. 

% Given a pair of images $I_A$ and $I_B$ and a face recognition system, our goal is to produce two saliency maps $S_A$ and $S_B$

\subsection{Similarity-based Randomized Input Sampling for Explanation (S-RISE)}
The previous section gives a new definition to the problem of explainable face recognition that an explanation method should both produce proper saliency maps for critical regions of the input image and interpret the decision of the recognition model. This study proposes a new model-agnostic explanation method for face recognition following the new definition of XFR. 

Despite being useful in principle, existing explainability tools for other image understanding tasks cannot be directly applied to the face recognition task. For example, the Randomized Input Sampling for Explanation (RISE) method explains a classification model by leveraging the categorical output probability of the classifier as the weight to aggregate the final saliency map. However, the decision-making process of a face recognition system mainly involves deep face representation extraction and similarity calculation between at least two images. To address this issue, this paper proposes a Similarity-based RISE algorithm (S-RISE) that leverages the similarity score as weights for the masks and provides explanation saliency maps without accessing the internal architecture or gradients of the face recognition system. 

Figure \ref{fig:srise} depicts an overview of the proposed S-RISE algorithm. In general, given a pair of images $\{img_A,$ $img_B\}$, a mask generator will first randomly produce a fixed number of masks. For each mask $M_i$, it will be applied to the input image, e.g. $img_A$. The masked $img_A$ and unmasked $img_B$ are then fed into the face recognition model respectively to capture the deep face representation. Afterward, the cosine similarity is computed as the weight of the corresponding mask. After iterating all the masks, the final saliency map $S_A$ for $img_A$ is the weighted combination of the generated masks.  
More specifically, the proposed S-RISE algorithm comprises the following two pivot steps.

\subsubsection{Mask Generation}
The conventional RISE algorithm for image classification tasks aims at generating multiple random non-binary masks. Basically, authors propose to first sample small binary masks and then upsample them to a larger resolution with bilinear interpolation, after which the masks have values between [0,1]. 
here we propose to simplify the process by randomly generating multiple small Gaussian-distributed masks in different locations. In practice, the mask generation process is as follows.

\begin{enumerate}
    \itemsep0.25em 
    \item [1.] Initialize the parameters of the mask generator, i.e. the total number of masks $N$, and the number and size of the Gaussian kernels in each mask.
    \item [2.] Sample multiple Gaussian kernels of fixed size in random locations.
    \item [3.] Merge all the generated Gaussian kernels into one mask
    \item [4.] Repeat step 2 and 3 to get $N$ different masks
\end{enumerate}

\subsubsection{Similarity-based Saliency Map Generation}

This subsection describes in more details how the S-RISE algorithm produces saliency maps for a triplet input. The S-RISE method is designed to explain the predictions between every independent pair of images due to the special decision-making process of the face recognition system. Thus, the triplet will be first divided into two groups, i.e. matching pair $\{I_p, I_m\}$ and non-matching pair $\{I_p, I_n\}$, and the S-RISE algorithm will explain their corresponding predictions, respectively. 
The following operation describes in detail the procedure used in the proposed S-RISE algorithm. 

\begin{enumerate}
    \itemsep0.25em 
    \item [1.] Split the triplet input into $\{I_p, I_m\}$ and $\{I_p, I_n\}$ pairs. 
    \item [2.] Sample $N$ masks, $M=\{M_i, 1\le i \le N\}$, using the mask generator. Apply $M_i$ to the $I_p$ and $I_m$ separately.
    \item [3.] Iterate $i$ from 1 to $N$: 
    \begin{enumerate}
    	\item [3.1.] Forward the masked probe image $I_p\odot M_i$ and unmasked mate image $I_m$ into the network for feature extraction, calculate similarity score $s^p_i$ between the deep features.
        \item [3.3.] Forward the masked mate image $I_m\odot M_i$ and unmasked probe image $I_p$ into the network for feature extraction, calculate similarity score $s^m_i$ between the deep features.
     \end{enumerate}
    \item [4.] Compute the weighted sum of masks $M_i$ with respect to the calculated similarity score $s^p_i$ to obtain a saliency map for the probe image $H_p = \sum^N_{i=1} s^p_i M_i$.
    \item [5.] Compute the weighted sum of masks $M_i$ with respect to the calculated similarity score $s^m_i$ to obtain a saliency map for the mate image $H_m = \sum^N_{i=1} s^m_i M_i$
    \item [6.] Repeat steps 2-5 for input pair $\{I_p, I_n\}$
    \item [7.] Re-weight the saliency maps for the matching pair and non-matching pair according to their similarity difference. 
\end{enumerate}

The mask generator randomly samples a fixed number of masks first for the matching pair and the S-RISE algorithm starts the iteration to compute the saliency map. 
The same steps are repeated for the non-matching pair. At the end, to illustrate a visually explainable heatmap, the values of all the saliency maps will be normalized and re-weighted based on the similarity difference between the matching and non-matching pairs.
% Notably, the masks are not applied to both input images at the same time and the saliency maps are produced separately because the similar regions of the two images are not necessarily in the same location. 

% Considering that the similar regions of the two images are not necessarily in the same location, the masks are not applied to both input images at the same time. 
% because the decision-making process of the recognition system is independent for both pairs. 
% Although the proposed explanation framework focuses on a triplet set, the S-RISE algorithm itself each time generates the saliency map explanation for an image pair. 
% It is notable that in each iteration of the S-RISE algorithm, the masks are not applied to both input images at the same time.  
% and starts the iteration of running the S-RISE algorithm. 
% Instead, the same mask is applied to the fir
% In the last step. 

\subsection{Evaluation Methodology}
The importance of rigorous evaluation methodologies has been overlooked in the field of explainable artificial intelligence and only a few metrics have been designed for visual saliency explanation methods. In the context of explainable image classification tasks, Petsiuk et al. \cite{petsiuk2018rise} insert or delete salient pixels from the input image and measure the change in the output classification probability. \cite{hu2022x} adopted the same ``Deletion'' and ``Insertion'' metrics to image retrieval task. In face recognition, \cite{castanon2018visualizing} quantifies the visualized discriminative features by playing a "hiding game", which iteratively obscures the least important pixels in the image sorted according to a produced attention map. 

In this paper we propose an evaluation approach to systematically validate the reliability and measure the performance of the general saliency-based XFR method. The conventional ``Deletion'' and ``Insertion'' evaluation metrics are adapted and improved to better fit the explainable face recognition framework. 
The main insight is that the explanation saliency map is expected to precisely highlight the most important regions of the face with the smallest number of pixels for the face recognition model to take a correct decision. 

In general, the Deletion and Insertion metrics measure how fast the similarity between two faces drops/rises to a threshold value after removing/adding saliency pixels from them. More specifically, the deletion process starts with original images, and the pixels with the highest saliency values are sequentially removed and replaced with a constant value. After removing each pixel, the similarity score is re-calculated until it is lower than a predefined threshold. On the contrary, the insertion process starts with the constant value, and the most critical pixels in the image sorted by the saliency map are added to the plain image. The similarity score is re-calculated each time after adding one pixel until it is larger than the threshold. The number of pixels deleted from or added to the image is accumulated until the recognition model changes the decision. Overall, the deletion and insertion metrics are defined as $\frac{\#Removed\text{ }pixels}{\#All\text{ }pixels}$ and $\frac{\#Added\text{ }pixels}{\#All\text{ }pixels}$. 
In practice, removing pixels from an image alters the original distribution and can eventually affect recognition results. Hence, the constant value above is set as the mean value of the specific image. 
% The threshold is the same value as what the face recognition model set to make the final verification. 

\section{Experimental Results}
\label{results}

\subsection{Implementation Details}
\subsubsection{Face Recognition Model Setup}
This paper utilizes the popular ArcFace \cite{deng2019arcface} face recognition approach, with a backbone network of ResNet-50 \cite{he2016deep}, for the explanation experiments. The architecture of the face recognition model remains unchanged during the explanation. The face recognition model is trained with the cleaned version of MS1M dataset \cite{deng2019arcface}, which is composed of approximately 5.1M face images belonging to 93K identities.

\subsubsection{Explanation Model Setup}
The proposed S-RISE explainability method does not require any training or access to the internal architecture of the face recognition model. The number of iterations in the explanation process is set to 1000 by default. 

\subsubsection{Dataset}
To validate the effectiveness of the proposed explainability approach, experiments have been carried out on testing images from several popular databases. First, a small subset of LFW \cite{huang2008labeled} is sampled, which comprises 50 triplets of faces. This subset is also used for the quantitative evaluation of the S-RISE method with the proposed metrics. 
In addition, some triplets are manually selected from CPLFW \cite{zheng2018cross} and Webface-Occ \cite{huang2021face} dataset for particular scenarios.

\subsubsection{Preprocessing}
All the test images strictly follow the same preprocessing steps as the training data. The MTCNN \cite{zhang2016joint} is first applied to detect faces and landmark points. All  images are then cropped, aligned, and resized to the size of 112x112 pixels.

\newcommand\w{\linewidth}
\newcommand\y{\linewidth}
%%%%%%%%%%%%%%%%%%%%%%% Sanity Check %%%%%%%%%%%%

\begin{figure}[t]
	\centering
	\begin{adjustbox}{width=\linewidth}
    \includegraphics[]{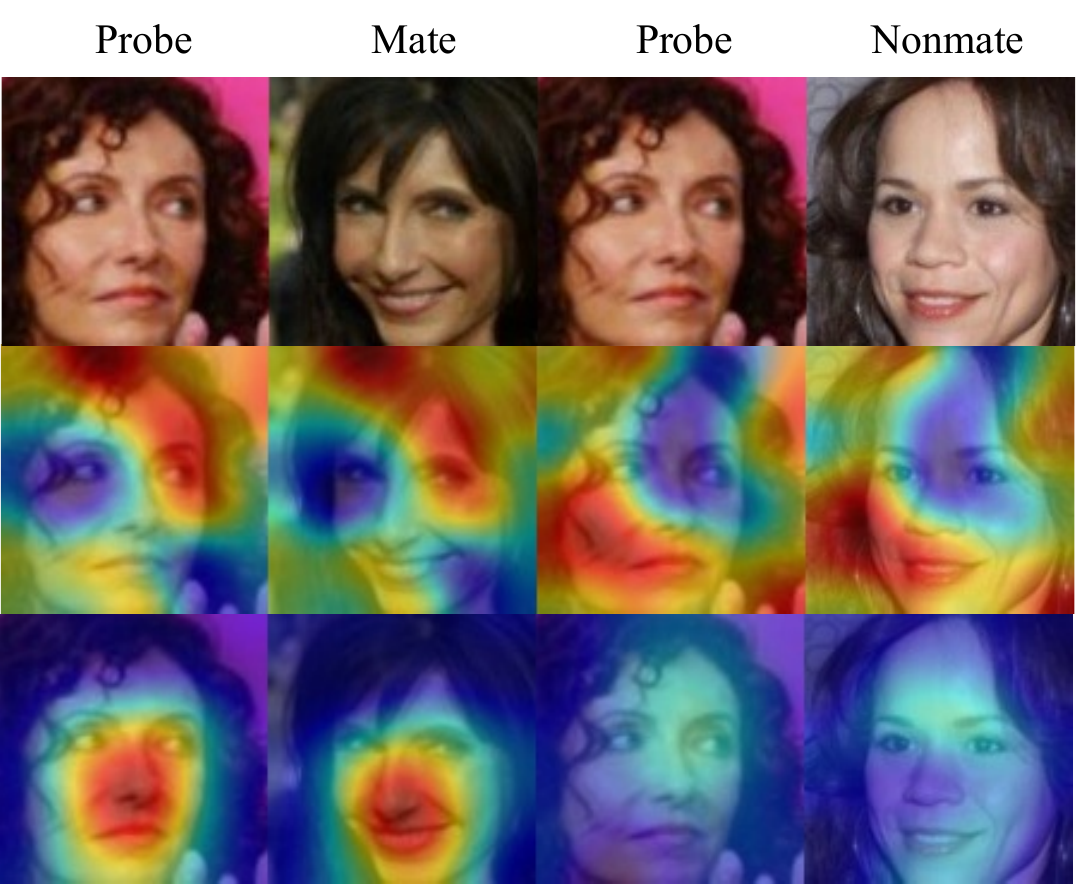}
	\end{adjustbox}
\caption{Sanity check for the S-RISE explanation method. The second row is the generated explanation heatmap for a CNN model with randomized parameters, while the third row is for a normal face recognition system. }
\label{fig:sanity}
\end{figure}

% \begin{figure}[h]
% \centering
% \begin{subfigure}[b]{\w}
%   % include first image
%   \includegraphics[width=\y]{Figures/SanityCheck/originaltriplet.png}  
% %   \label{fig:sub-first}
% \end{subfigure}%
% \hfill\begin{subfigure}[b]{\w}
%   % include first image
%   \includegraphics[width=\y]{Figures/SanityCheck/sanitycheck.png}  
% %   \label{fig:sub-first}
% \end{subfigure}%
% \hfill
% \begin{subfigure}[b]{\w}
%   \includegraphics[width=\y]{Figures/SanityCheck/sanitycheck2.png}  
% %   \label{fig:sub-first}
% \end{subfigure}%
% \caption{Sanity check for the S-RISE explanation method. The second row is the generated explanation heatmap for a CNN model with randomized parameters, while the third row is for a normal face recognition system. }
% \label{fig:sanity}
% \end{figure}
%%%%%%%%%%%%%%%%%%%%%%%%%%%%%%%%%%%%%%%%%%%%%%%%%%%%%%%%%%%%%%%%%%%%%

\subsection{Sanity Check for the Proposed Explanation Method}
A recent study \cite{adebayo2018sanity} %Sanity Checks for Saliency Maps
has questioned the validity of saliency methods that some of the produced explanation heatmaps can be independent both of the model and of the data generating process. They propose a model parameter randomization test for a sanity check, where the weights of a deep neural network are randomly initialized before applying the explanation method. 
In the context of this paper, a saliency method can also provide some visually compelling results by simply applying a heatmap that concentrates on the center of the face and claim it to be an attention map without analyzing the behavior of the recognition model. Therefore, a similar idea for the sanity check is adopted by directly loading irrelevant ResNet parameters tuned for other vision tasks.

The explanation results in the second row of Figure \ref{fig:sanity} show that the random parameters of the deep model will result in nonsense saliency maps, which validates that the proposed S-RISE algorithm fully relies on the trained recognition model and is capable of producing meaningful interpretations. 

% employs the same model parameter randomization test as proposed in \cite{} for a sanity check, 
% where the weights of the recognition model are randomly initialized before applying the explanation method.

%%%%%%%%%%%%%%%%%%%%%%% Example figures for visual explanation results %%%%%%%%%%%%

\begin{figure}[t]
	\centering
	\begin{adjustbox}{width=\linewidth}
    \includegraphics[]{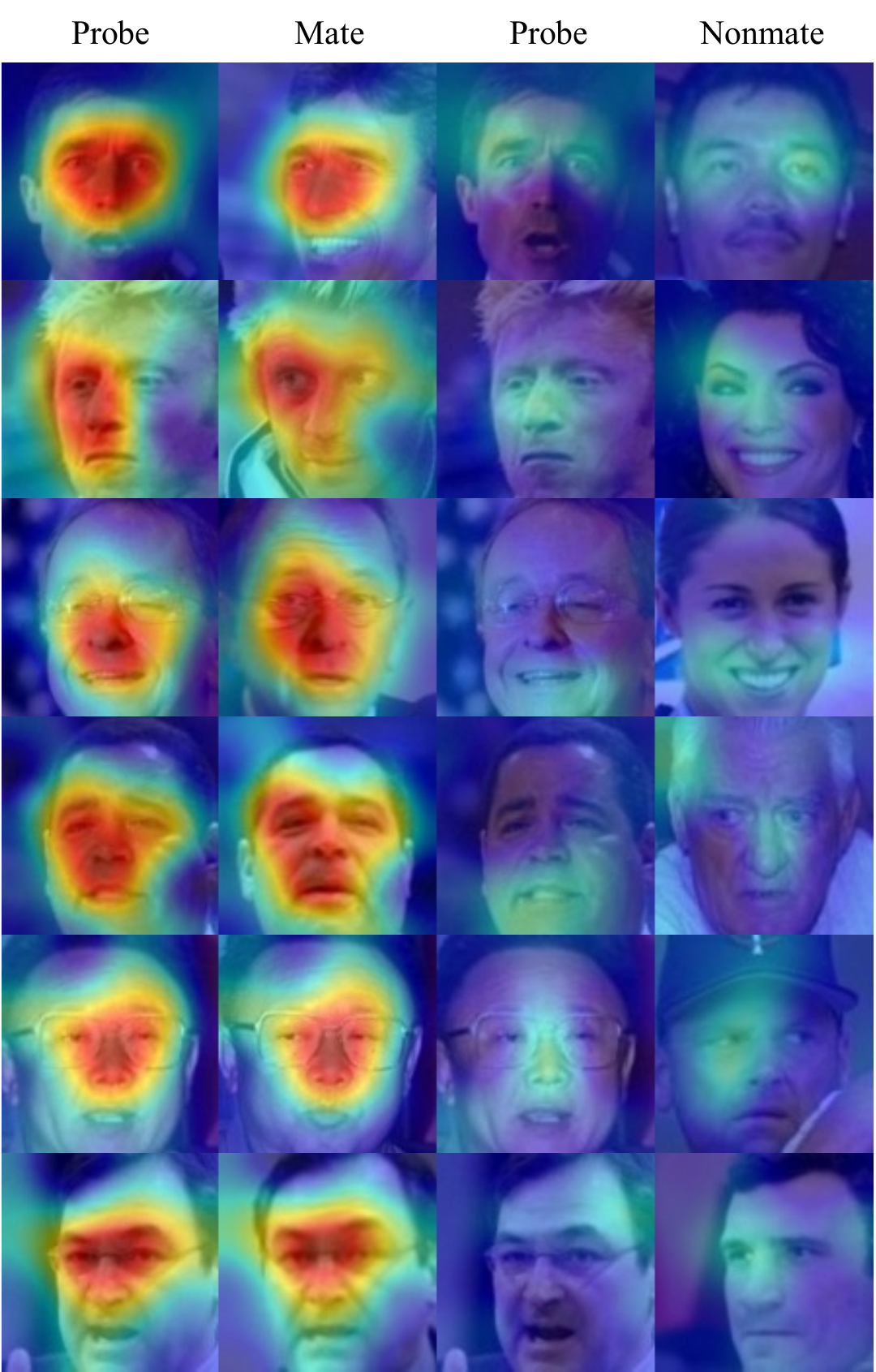}
	\end{adjustbox}
    \caption{Saliency map explanations for the FR model's prediction on the matching (left) and non-matching (right) image pairs.}
    \label{fig:highconf}
\end{figure}

\begin{figure}[t]
	\centering
	\begin{adjustbox}{width=\linewidth}
    \includegraphics[]{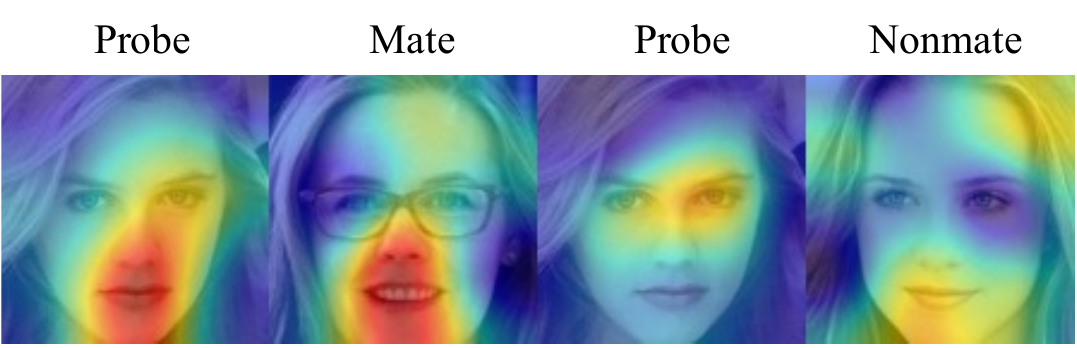}
	\end{adjustbox}
    \caption{Saliency map explanation on failed predictions of the face recognition model.}
    \label{fig:lowconf}
\end{figure}

% \begin{figure}[t]
% \centering
% \begin{subfigure}[b]{\w}
%   % include first image
%   \includegraphics[width=\y]{Figures/Difficult/Alicia_Sliverstone.png}  
% %   \label{fig:sub-first}
% \end{subfigure}%
% % \hfill
% % \begin{subfigure}[b]{\w}
% %   \includegraphics[width=\y]{Figures/Difficult/Alan_Trammell.png}  
% % \end{subfigure}%

% \caption{Explanation on failed predictions of the face recognition system.}
% \label{fig:lowconf}
% \end{figure}
%%%%%%%%%%%%%%%%%%%%%%%%%%%%%%%%%%%%%%%%%%%%%%%%%%%%%%%%%%%%%%%%%%%%%

%%%%%%%%%%%%%%%%%%%%%%% Example figures for Self-Occlusion %%%%%%%%%%%%
\begin{figure}[t]
	\centering
	\begin{adjustbox}{width=\linewidth}
    \includegraphics[]{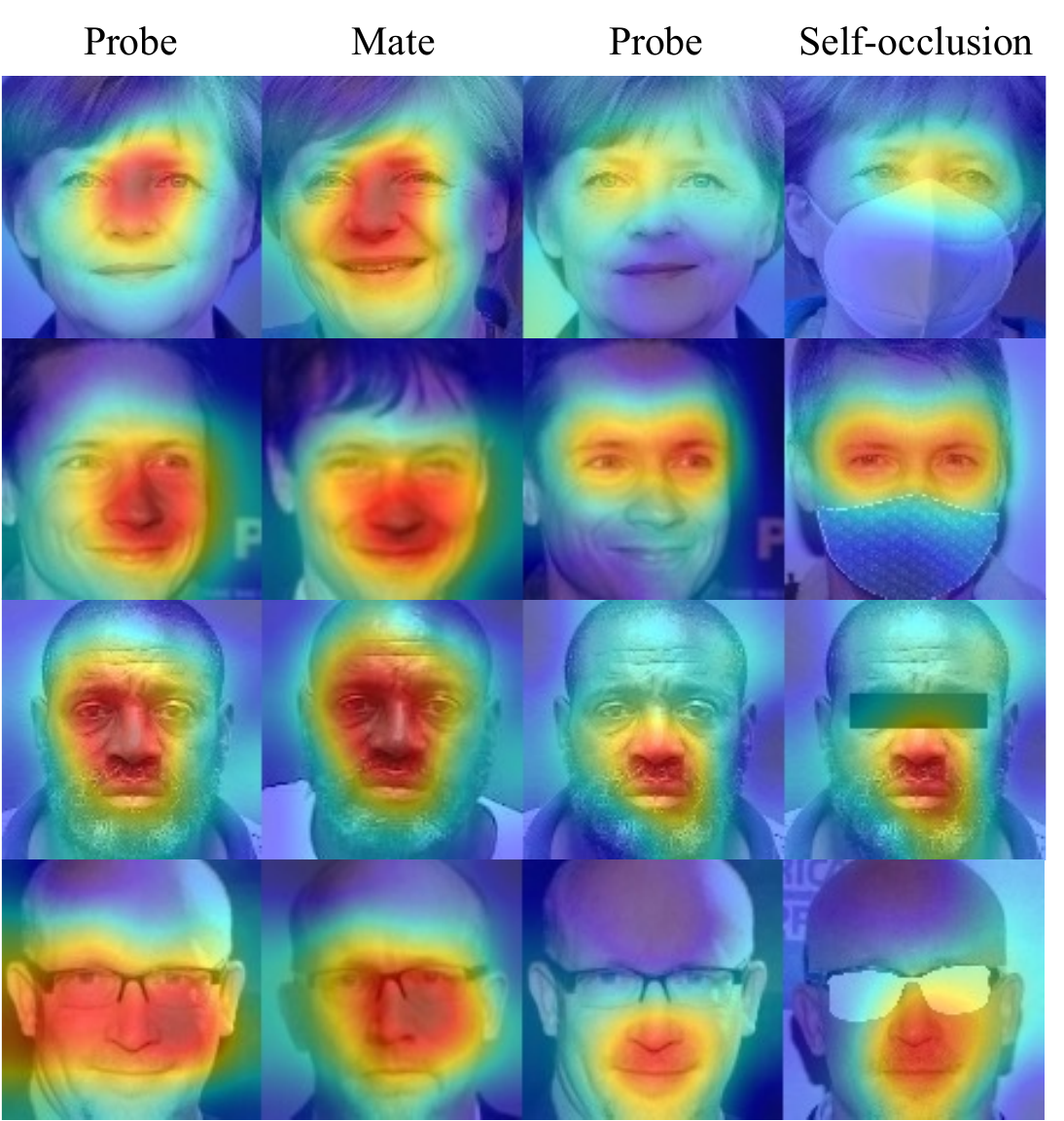}
	\end{adjustbox}
    \caption{Saliency map explanations for the predictions of the FR model on partially-occluded faces.}
    \label{fig:occlusion}
\end{figure}

% \begin{figure}[t]
% \centering
% \begin{subfigure}[b]{\w}
%   % include first image
%   \includegraphics[width=\y]{Figures/Occlusion/Merkel.png}  
% %   \label{fig:sub-first}
% \end{subfigure}%
% \hfill
% \begin{subfigure}[b]{\w}
%   \includegraphics[width=\y]{Figures/Occlusion/0_494340.png}  
% %   \label{fig:sub-first}
% \end{subfigure}%
% \hfill
% \begin{subfigure}[b]{\w}
%   \includegraphics[width=\y]{Figures/Occlusion/Prison.png}  
% %   \label{fig:sub-first}
% \end{subfigure}%
% \hfill
% \begin{subfigure}[b]{\w}
%   \includegraphics[width=\y]{Figures/Occlusion/0_493213.png}  
% %   \label{fig:sub-first}
% \end{subfigure}%
% \hfill
% \begin{subfigure}[b]{\w}
%   \includegraphics[width=\y]{Figures/Occlusion/0_494791.png}  
% %   \label{fig:sub-first}
% \end{subfigure}%
% \hfill
% \begin{subfigure}[b]{\w}
%   \includegraphics[width=\y]{Figures/Occlusion/0_499104.png}  
% %   \label{fig:sub-first}
% \end{subfigure}%
% \caption{.}
% \label{fig:occlusion}
% \end{figure}
%%%%%%%%%%%%%%%%%%%%%%%%%%%%%%%%%%%%%%%%%%%%%%%%%%%%%%%%%%%%%%%%%%%%%

% In real-world deployments, a face recognition system can make correct prediction with either high or low confidence

\subsection{Visual Results of Saliency Maps}
This section presents the visual results of the saliency map generated by our proposed S-RISE algorithm. The results are divided into two groups. The first group shows the visual explanations for the predictions that the face recognition model correctly makes with high confidence, see Figure \ref{fig:highconf}. As a result, the produced saliency map properly highlights the regions between the matching pairs that the FR model believes are very similar. As for the probe and nonmate pairs, the heatmaps also represent similar regions but they are much shallower, indicating low similarities between them, which explains why the model believes they are not from the same subject.
% As a result, the produced saliency map properly highlights the regions that are similar to the FR model between the matching pairs. As for probe and nonmate pairs, the heatmaps also represent similar regions but they are much shallower, indicating low similarities between them.

While the current deep face recognition model generally achieves high prediction accuracy with high confidence in most standard scenarios, it mistakenly identifies two subjects as the same person in some cases. The second group of results consists of a manually selected triplet that the FR model fails to differentiate, see Figure \ref{fig:lowconf}. 
According to the explanation heatmap, although it recognizes the probe-mate pair with high confidence, it also allocates relatively high importance to the eye and mouth regions to the probe-nonmate pairs, which explains why the FR model fails the verification.

% The saliency maps for two example triplets in Figure \ref{fig:lowconf} show that the explanation model allocates relatively high importance to the eye and mouse regions to the probe-nonmate pairs, which explains why the model mistakenly verifies them as matching.

% that the FR model mistakenly verifies the probe and nonmate as the same person. 
% The heatmaps gives much higher saliency values to the eye and mouth regions of the probe and nonmate images, which explains why the model 

% Figure \ref{fig:lowconf} displays the explanation results of the 
% can fail or make correct decisions with low confidence in some cases. For example, 
% Figure \ref{} displays the explanation saliency map of the decisions with low confidence. 

To further validate the effectiveness of the proposed explainability model, an additional test has been performed with self-occluded faces. Studies \cite{lu2022novel} have shown that the current deep face recognition model is capable of identifying partially occluded faces despite lower confidence. In this case, an ideal explanation method should provide low saliency values for occluded pixels while high values for other similar regions. 
In this experiment, the non-matching face in a triplet is replaced by an occluded image from the same subject as the probe face. 
As presented in Figure \ref{fig:occlusion}, the S-RISE algorithm explains that the FR model manages to verify them through the eye regions when images are occluded by facial masks, and through mouth and nose areas when masked by sunglasses. 

%%%%%%%%%%%%%%%%%%%%%%%%%%%%%%%%%%%%%%%%%%%%%%%%%%%%%%%%%%%%

\begin{table}[t]
  \centering
  \caption{Quantitative evaluation of saliency maps using proposed Deletion and Insertion metrics.}
    \begin{tabular}{c|c|ccc}
    \toprule
    Methods & Iterations & Deletion & Insertion & Average \\
    \midrule
    \multirow{3}{*}{S-RISE} & 10 & 0.4466 & 0.3582 & 0.4024 \\
     & 100  & 0.2617 & 0.1983  & 0.2300 \\
     & 500  & 0.2071 & 0.1459  & 0.1765 \\
     & 1000 & 0.2077 & 0.1384 & 0.1731 \\
    \bottomrule
    \end{tabular}%
  \label{tab:eval}%
\end{table}%

\begin{figure}[t]
	\centering
	\begin{adjustbox}{width=\linewidth}
    \includegraphics[]{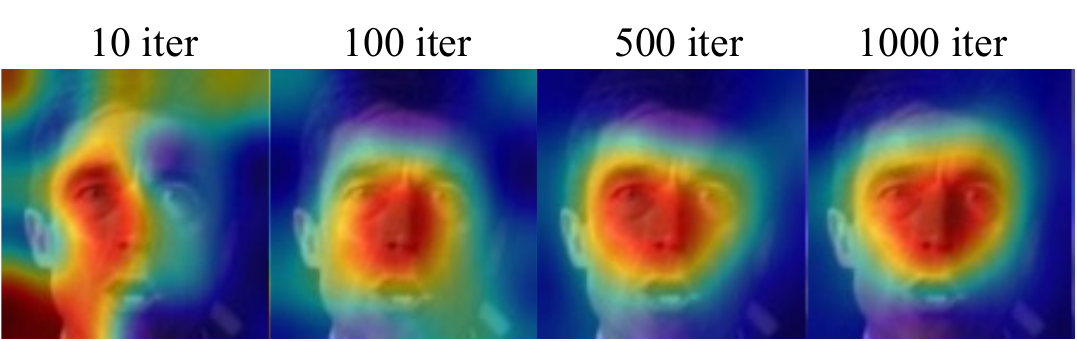}
	\end{adjustbox}
    \caption{Saliency map generated by S-RISE algorithm with different iteration configurations.}
    \label{fig:diffiter}
\end{figure}
%%%%%%%%%%%%%%%%%%%%%%%%%%%%%%%%%%%%%%%%%%%%%%%%%%%%%%%%%%%%

\subsection{Quantitative Evaluation}
This section reports the deletion and insertion metrics as quantitative evaluations for the proposed S-RISE explanation model. The experiments are conducted on a small subset of the LFW dataset. In general, the metrics measure the percentage of modified pixels in order to change the decision of the FR model, and the smaller, the more accurate the explanation saliency map. 

Table \ref{tab:eval} shows the quantitative evaluation for the proposed S-RISE algorithm under different configurations in terms of iterations. The metrics show that a small number of iterations results in poor explanation performance. On the other hand, the metrics gradually converge at around 1000 iterations, which corresponds to stable and accurate saliency maps. Figure \ref{fig:diffiter} further validates the conclusion drawn by quantitative evaluation results. 
% The metrics are interpreted as follows. After deleting around 20.8\% of pixels that are accompanied by the highest saliency values, the model fails to recognize the subject. After inserting around 13.8\% of the most critical pixels into a plain image, the recognition model manages to identify the subject.  

% According to the produced saliency map,
% 
\section{Conclusion}
\label{conclusion}

In this paper, a new explainable face recognition framework is conceived. The proposed S-RISE algorithm is capable of producing insightful saliency maps to interpret the decision of a deep face recognition system. Extensive visual results of saliency maps have demonstrated the effectiveness of the method. Furthermore, two novel evaluation metrics are proposed in this work to measure the quality of the saliency maps generated by the S-RISE algorithm. In the future, the evaluation method will serve as a public benchmark for general visual saliency map-based XFR methods.

%%%%%%%%%%%%%%%%%%%%%%%% Acknowledgement %%%%%%%%%%%%%%%%%%%%%%%%%%%%
\begin{acks}
The authors acknowledge support from CHIST-ERA project XAIface (CHIST-ERA-19-XAI-011) with funding from the Swiss National Science Foundation (SNSF) under grant number 20CH21 195532.
\end{acks}

%%%%%%%%%%%%%%%%%%%%%%%% References %%%%%%%%%%%%%%%%%%%%%%%%
\bibliographystyle{ACM-Reference-Format}
\bibliography{ref}

% \appendix


\end{document}